# Complexity

# An adaptive data-driven approach to solve real-world vehicle routing problems in logistics

## Emir Žunić [1,2] • Dženana Đonko [1] • Emir Buza [1]

1    Faculty of Electrical Engineering, University of Sarajevo, Bosnia and Herzegovina

2    Info Studio d.o.o. Sarajevo, Bosnia and Herzegovina

Correspondence should be addressed to Emir Žunić
emir.zunic@etf.unsa.ba; emirzunic86@hotmail.com; emir.zunic@infostudio.ba
ORCID: 0000-0003-1950-7816
Google Scholar: https://scholar.google.com/citations?user=FN3kxnQAAAAJ&hl=hr
LinkedIn: https://www.linkedin.com/in/emir-zunic/
Tel: +387 61 858 695

## Abstract

Transportation occupies one-third of the amount in the logistics costs, and accordingly transportation systems largely influence the performance of the logistics system. This work presents an adaptive data-driven innovative modular approach for solving the real-world Vehicle Routing Problems (VRP) in the field of logistics. The work consists of two basic units: (i) an innovative multi-step algorithm for successful and entirely feasible solving of the VRP problems in logistics, (ii) an adaptive approach for adjusting and setting up parameters and constants of the proposed algorithm. The proposed algorithm combines several data transformation approaches, heuristics and Tabu search. Moreover, as the performance of the algorithm depends on the set of control parameters and constants, a predictive model that adaptively adjusts these parameters and constants according to historical data is proposed. A comparison of the acquired results has been made using the Decision Support System with predictive models: Generalized Linear Models (GLM) and Support Vector Machine (SVM). The algorithm, along with the control parameters, which using the prediction method were acquired, was incorporated into a web-based enterprise system, which is in use in several big distribution companies in Bosnia and Herzegovina. The results of the proposed algorithm were compared with a set of benchmark instances and validated over real benchmark instances as well. The successful feasibility of the given routes, in a real environment, is also presented.

# 1 Introduction

Since logistics advanced in the 1950s [1], numerous studies were carried out within various application domains. Due to the trend of nationalisation and globalisation in recent decades, the importance of logistics management has been growing in various domains. For industries, logistics help to optimise the existing production and distribution processes based on the same resources through management techniques for promoting the efficiency and competitiveness of enterprises. The essential element in a logistics chain is the transportation system, which connects separated activities. Transportation occupies one-third of the amount in the logistics costs, and transportation systems largely influence the performance of the logistics system. Transportation is required over within the whole production procedure, starting from manufacture, delivery to the end consumer, and even return of goods. Only an excellent coordination between each component would maximize the benefits. Without well-developed transportation systems, logistics could not bring its advantages into full play. In logistics, such transportation system could provide better efficiency, reduce operation costs, and promote service quality [2]. The successful solving of vehicle routing problems can significantly improve company's operations in the field of transportation.

The vehicle routing problem is a generalization of the Traveling Salesman Problem (TSP), which is one of the most studied optimization problems. The problem is concerned with a travelling salesman who has a task to visit a set of cities in the shortest possible path, having that each city is visited only once, and that the starting city must be the finishing city as well. It is necessary to find the shortest route, which fulfils the previous condition [3]-[5]. When the previously defined problem is concerned with more than one travelling salesman, having that all of them are in the same city, then the Vehicle Routing Problem at its starting state is defined. It is necessary to find a set of shortest possible routes for the travelling salesmen, such that each city is visited by only one salesman. This variant of the problem is called the Multiple Travelling Salesman Problem – MTSP [6]. MTSP is similar to the basic form of the VRP. There are two main reasons why VRP is one of the most studied problems in the academic community. Those are two characteristics which VRP has, and which TSP lacks [7]. Firstly, there are still no algorithms which are as successful in solving the VRP problem as the most successful algorithms for solving the TSP problem are. Secondly, VRP is a problem set exceptionally applicable in practice. Various difficulties occurring in companies that deal with transport and logistics formulated with different variants of the VRP.

To get the entire effect in practice, the approach and model (algorithm) presented in this paper for solving the real-world VRP should be adequately applied and validated in real conditions. Many facts indicate that to use this approach in practice, in the area of freight and logistics, it is important to consider various factors dependent on many parameters, primarily the number of served customers, and whether the sold/delivered goods are packages or pieces. This directly affects the way of storing the goods and loading the vehicles, through various natural limitations, such as the fact that some customers often have to use the same vehicle due to the temperature or other conditions; the realistic duration of unloading the goods at the delivery points; calculating the costs of transport routes; dependence on the loading and vehicle types; legal limits on the maximum service duration of a vehicle and a driver, and others. The human mind, and therefore the transport managers in companies in real-world situations create transportation routes by forming independent clusters to which vehicles are joined, from the available vehicle fleet. However, this approach which is based on regions cannot take into the consideration all the mentioned factors when creating the transportation routes, in a way that satisfies all the constraints. Most of the available software solutions operate on the same principle, i.e., grouping the customers into the clusters (regions), having the possibility to further bind multiple adjacent regions that lie in the same route. This clustering approach is often unable to solve complex problems that occur in practical applications in logistics, resulting in the unfeasible routes. It does not provide the best solution, especially in the cases of larger cities, where it is extremely difficult to define logical and completely separated customers' regions. Clustering approach also falls into the performance issues, caused by many delivery points, an extremely heterogeneous vehicle fleet, and a lot of different constraints that need to be fulfilled. Therefore, based on the mentioned facts, the purpose of this work is divided into two parts: (i) solving the complex real-world VRP in logistics using the proposed innovative multi-step algorithm, while meeting all of the appointed realistic constraints, (ii) adjusting the parameters and constants of the given model and algorithm by using the historical data, on an adaptive way. The proposed algorithm for solving the complex real-world VRP is based on the principle of penalization, and as such, uses numerous constants and parameters that are additionally adjusted using the historical data. Hence, this approach is "data-driven". Transportation routes which are the final result of these two interconnected entities, are mostly feasible from the practical point of view, which is the most important point for any company that wants to successfully create the best possible transportation route. Therefore, this paper presents the application of a new innovative approach and concept for solving VRP using the real data of one of the largest distribution companies in Bosnia and Herzegovina. Dataset is public and stored at 4TU.ResearchData and available for use to other researchers, as the new benchmark data ([8]-[9]). The validation of this approach was first done upon the standardized benchmark data, then upon the actual routes of the mentioned distribution company.

In more details, in this work we present and propose a multi-step algorithm that solves the real-world VPR. The algorithm implements four successively connected steps, comprising several data transformation methods, heuristics, and the Tabu search. The main goal of the proposed algorithm is to solve the real-world VRP problems in logistics with minimal cost, while meeting all of the realistic constraints. The proposed algorithm consists of a number of control parameters and constants, and hereby, we also present the prediction model for adaptively setting up and adjusting the parameters according to the historical data. In this way the suggested approach of solving the VRP problem acquires a more adaptive character. A comparison of the acquired result was made using the realized decision support system for the analysed prediction models: The Generalized Linear Models (GLM) and Support Vector Machine (SVM). A comparison has been made between the results of the proposed algorithm with a set of benchmark instances, also results over real benchmark data have been presented which also contain additional realistic information and constraints. The proposed model and concept, along with the managing parameters which were acquired using the proposed prediction method, was incorporated into the web-based enterprise system which is in productional use in the real sector. In the example of the distribution company and its realistic data used in this paper, all the transportation routes obtained by the proposed approach in the testing period lasting for three months, were completely feasible, respecting all of the restrictions and constraints. The financial savings obtained by using these routes are considerable in the area of transportation of the given company, and in this way, the proposed approach is successfully validated in practice.

Based on a detailed overview of the state-of-the-art which is presented in the next section (Section 2), and analysis of real problems of optimizing transport routes, an area for proposing a new, modular approach and algorithm for adaptive solving of the complex VRP problems with realistic constraints, has been observed. The third section (3.1) contains a detailed description of the proposed multi-step algorithm, which consists of several steps which are incorporated in a single unique unit. The proposed algorithm for solving the VRP problem consists of its constants and control parameters. The third section (3.2) also presents an innovative approach for adjusting the given parameters. Discussion of the results has first been done on standardized input datasets, and afterwards on an real dataset of one of the biggest distribution company in Bosnia and Herzegovina. The given data has been available to other researchers as well. At the end, results of part of the system responsible for adjusting the control parameters, have been discussed. All of the aforementioned discussions of results are described and explained in more detail in section four, as well as successful feasibility of the given routes in real-world environment. Section five presents conclusions of the work and guidelines for further research in this scientific field.

## 2 Related work

As stated in [10], the transportation of the product is a significant component of the total product cost (10%). The routing problem complexity increases exponentially when the number of customers or vehicles increases. Even for smaller instances, manual routing is becoming a difficult task for humans, and software based solutions can provide better performance than an experienced worker. Many algorithms for heuristic solving of the vehicle routing problem were described in the literature. Various approaches that use modern and recently introduced algorithms are described In [11] state of the art and problem review is given. The paper makes a reference to 277 articles that present different approaches to solving the vehicle routing problem published between 2009 and 2015.

The vehicle routing problem mostly includes various real-world constraints. All those constraints have to be taken into consideration a feasible solution to real-world usage.

The most famous constraints are time windows. The time window represents the time interval when the customer can be served. In paper [12] authors presented a mathematical model for the vehicle routing problem with access time windows, a version of the VRP suitable for planning delivery routes in a city with accessibility restriction, which bans the access of freight vehicles to central urban areas in many European cities. They used the model to find exact solutions to small problem instances based on a case study and then compare the performance over larger instances of a modified savings algorithm, a genetic algorithm, and a Tabu search procedure. The results do not show a clear prevalence of any of them, but confirming the significance of those additional costs and externalities. In [13] the adaptive memetic algorithm for minimizing distance in vehicle routing problem with time windows (VRPTW) is described. In [14], different metaheuristic approaches for solving VRPTW are described, such as Particle Swarm Optimization (PSO), Ant Colony Optimization (ACO) or Artificial Bee Colony (ABC). In [15] a hybrid of ACO and Firefly algorithm (FA) for solving vehicle routing problems is given. The VRPTW has been further tested. In [16], the improved simulated annealing algorithm is described.

The vehicle capacity is an important constraint used in real-world examples. In [17] improved K-nearest neighbour algorithm is used to solve capacitated VRP (CVRP). In [18] an improved simulated annealing for the CVRP is described. In [19], an *Evolutive Tabu-Search* approach is used for CVRP.

The number of depots varies for different companies. When more than one depot is in usage, the problem is called a multi-depot vehicle routing problem (MDVRP). In [20] an improved ACO algorithm is used for multi-depot green VRP. Enhanced differential evolution algorithms for solving MDVRP are used in [21]. In [22] discrete FA is used to solve asymmetric multi-depot VRP. In [23] other approaches and literature review for multi-depot vehicle routing problem is described.

One of the most important realistic constraints is fuel consumption, whether it is logistics of the own vehicle fleet or outsourced logistics. Outsourced logistics operation to third-party logistics has attracted more attention in the past several years. However, very few papers analysed fuel consumption model in the context of outsourcing logistics. In paper [24] authors presented a hybrid Tabu search algorithm for a real-world open vehicle routing problem involving fuel consumption constraints. Experiments in this paper were conducted on instances based on real road data of Beijing, China, considering that outsourced logistics plays an increasingly important role in China's freight transportation.

In literature and practice, many other constraints are considered, such as heterogeneous fleet VRP [25], city VRP [26], etc. In [27], a taxonomic review of the vehicle routing problem is given with different approaches and a methodology for classifying the literature. Often, the problems of transport optimization have a dynamic interpretation. Ride-sharing services are transforming urban mobility by providing timely and convenient transportation to anybody, anywhere, and anytime. However, most of the mathematical models do not fully address the potential of ride-sharing. Authors in paper [28] presented a more general mathematical model for real-time high-capacity ride-sharing that (i) scales to large numbers of passengers and trips, and (ii) dynamically generates optimal routes with respect to online demand and vehicle locations. The algorithm applied to fleets of autonomous vehicles and also incorporates rebalancing of idling vehicles to areas of high demand. This framework is general and can be used for many real-time multivehicles, multitask assignment problems.

From the aspect of practical application in logistics, it is always important to do all the necessary pre-treatments which lead to the optimal transportation routes, satisfying all the constraints. In paper [29], the authors presented a multi-phase hybrid approach with clustering, dynamic programming, and a heuristic algorithm to solve a collaborative multiple-centre vehicle routing problem (CMCVRP). CMCVRP is a multi-constraint combinatorial and game optimization issue containing both vehicle routing optimization and profit distribution procedures. The CMCVRP is generally used to study the logistics network structure adjustment from a non-optimal network structure to a collaborative multiple DCs network optimization structure. The optimization of CMCVR can effectively improve vehicle loading rate and reduce the crisscross transportation phenomenon. Designing a reasonable profit distribution mechanism is a critical step in CMCVR optimization. Collaboration can be organized through a negotiation process by a logistics service provider. On the other hand, the study [30] establishes a linear optimization model to minimize the total cost of a two-echelon logistics joint distribution network. An improved ant colony optimization algorithm integrated with the genetic algorithm is presented to serve customer clustering units and resolve the model formulation by assigning logistics facilities. Collaborative two-echelon logistics joint distribution network can be organized through a negotiation process via logistics service providers or participants existing in the logistics system, which can effectively reduce the crisscross transportation phenomenon and improve the efficiency of the urban freight transportation system.

As it can be observed, all heuristic algorithms contain parameters that must be set for the algorithm to provide a quality and usable solution. This makes the parameter-setting problem a significant research area because the results of the algorithm significantly depend on the parameter values. In the majority of the cited research papers, a fact has been mentioned which indicates that every real-world VRP problem is a slightly bit different from the VRP problem that is the most similar to it. That is affected by two sets of parameters (control), among other things: (i) certain realistic constraints and input data constants, and (ii) constants of the used algorithm.

Each company that requires an implementation of the VRP has its constraints that are defined by the business policy of the given company. That is why the literature often states that constraints and restrictions in these types of problems are non-standard. Lee describes one of these problems in great detail in [31], along with a solution proposal on a concrete realized example. Data Mining (DM) techniques and methods are also often used for adjusting the realistic constraints of the VRP, as well as Machine Learning (ML) algorithms, and other methods such as *Fuzzy* logic or Neural Networks (NN). There are several available papers which describe the application of statistical methods for these causes. Some kinds of interesting real examples are presented in papers [32]-[33]. An especially interesting example of the classic application of real data is presented in research [34] where the term data-driven solving of the VRP problem is introduced. This work also mentions for the first time an additional phase, which can be used in a real environment, and that is the human-computer interaction phase, which enables that the end user has the possibility of manual processing and modification of the suggested routes. No matter how perfect the algorithm seems, there are always real situations that are impossible to predict and classify in advance, and that is why that possibility is needed in practical systems.

Each of the analysed approaches and algorithms for solving the VRP problem, including the one presented in this work, consists of certain constants and control parameters. Those parameters and constants are used to determine certain weight factors, punitive factors by individual criteria, depending on the importance of the criteria for the result of the real situation of vehicle routing and others. In literature, this approach is known as the *Parameter Setting Problem* (PSP). The most interesting work on this topic was presented by Calvet et al. [35] in which they describe a statistical approach for setting the parameters for metaheuristic algorithms, and which is applied to the VRP problems.

# 3   Proposed adaptive data-driven approach

The key element in the supply chain is the transportation system that unites different, spatially and temporally separated activities. The transportation includes one-third of logistics costs and significantly affects the performances of the logistic system. Distribution companies usually have problems where they are not able to optimize their transportation activities in the best possible way and therefore lose considerable financial resources. Two basic units are presented in this paper (Figure 1):

- a multi-step algorithm, that is able to optimally solve real-world and extremely complex VRP problems, satisfying most of the constraints that can occur in practice (section 3.1),
- a proposed algorithm consists of appropriate constants and parameters that can be set based on historical value, so the second part of this section (section 3.2) presents data-driven approach for setting the control parameters of the proposed algorithm.

This two-component model could be represented as a diagram seen in Figure 1. In addition to historical data, the other data are also taken into consideration, such as Global Positioning System (GPS) and Geographic Information System (GIS) data.

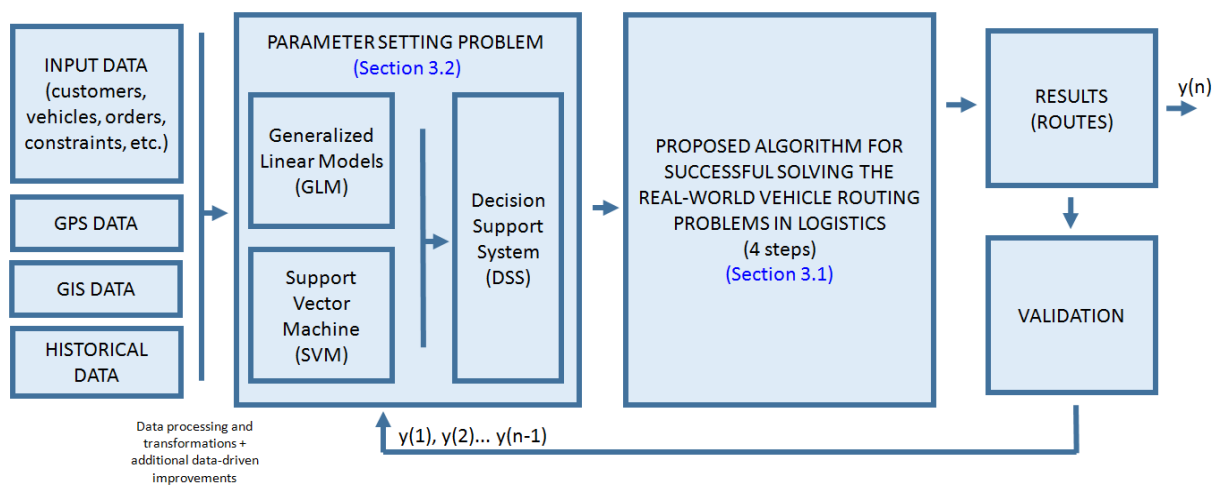

Figure 1: Proposed data-driven approach for adaptive solving the vehicle routing problem

Transportation routes obtained this way are optimal from the algorithmic point of view and completely feasible in a real environment, which is the most important fact, from the aspect of the practical application of such systems. The transportation costs are significantly reduced, routes are feasible, and customers of the company are more satisfied as their requirements are fulfilled by using this multi-component approach on the real example of the distribution company.

## 3.1 Multi-step algorithm for solving real-world VRP problems in logistics

The proposed algorithm that solves the heterogeneous fleet vehicle routing problem with time windows (HVRPTW), and satisfies realistic constraints, consists of four steps, or rather two steps, one intermediate step and one post-step (Pseudocode 1). In the first step, the initial solution to the problem is created, which is improved in later steps. A modification of the heuristic Clarke-Wright algorithm was implemented. After that, the second step (intermediate step) strives to decrease the number of routes. The solution which is the result of the second step serves as an initial solution to the local search based on Tabu search, or the third step. After Tabu search, the fourth step (post-step) optimizes the order of customers within each acquired route. Before that, the transformation of input data, time windows of customers and time distances is done, which enables the duration of unloading at each customer to become equal to 0 and simplifies the problem. The algorithm also solves the problem where some customers cannot be serviced by certain vehicles from the fleet (*Site-Dependent Vehicle Routing Problem* - SDVRP), which is one of the most commonly used realistic constraints in practice.

```
Pseudocode 1: Steps of the proposed algorithm

/* Pre-step (3.1.1) */
  ConnectingToTheDatabase (connection);
  DataInitialization(customersList, vehiclesList, depot, sdvrpConstraints);
  PreparationOfTheInputData(customersList, vehiclesList, depot, sdvrpConstraints);
  TimeWindowsTranspformations(customersList);
  RoutesInitialization(customersList, routes);
/* First step (3.1.2) and (3.1.3) */
// initial solution becomes optimal routes
  ClarkeWrightSavingsAlgorihm(routes, vehiclesList);
  AssigningVehiclesToRoutes(routes, vehiclesList);
/* Second step (3.1.4) */
  initialSolution = ReducingTheNumberOfRoutes(routes, customersList, vehiclesList);
/* Third step (3.1.5) */
// the initial solution marked as the initialSolution becomes the optimal routes
  TabuSearch(initialSolution, customersList, vehiclesList, sdvrpConstraints);
/* Fourth step (3.1.6) */
  RoutesOptimization();
  RoutesOptimization2Customers();
  RoutesOptimization3Customers();
  InverseTransformation(customersList, initialSolution);
/* Final step */
// the result of the fourth step is the optimal list of routes by vehicle is RouteResList
  StoreOptimalRoutesInTheDatabase(RouteResList);
```

Before the description of the proposed algorithm, the formulations of the basic concepts as well as the problem will be briefly introduced:

**Route** – this implies a row of customers and a vehicle which travels that route. The route has much more data and characteristics, but it is derived from these two, which means that the route is defined by a row of customers and one vehicle.

**Solution** – one solution is a set of routes where each customer is in exactly one route. The solution depends on the order of each customer in a transport route of a corresponding vehicle. The time of arrival at the customer is determined so that it is the earliest possible time (meaning that time is always equal to the start of the time window or it is equal to the time that was necessary to get to the previous customer). The time of arrival at the first customer is equal to the start of his time window, or it is equal to the start of the time window of the vehicle increased by the time necessary for the vehicle to arrive at the warehouse of the first customer.

**Realistic constraints** which are covered in the proposed algorithm and approach – a lot of the constraints which could be met in the real environment are included in the proposed four-steps algorithm and approach. Some of them are as follows: a large number of the customers with their time-windows included, heterogeneous fleet of vehicles, working hours of each vehicle, working hours of the drivers as well, working hours of the depot, site-dependent constraints, blocking roads for each vehicle in the fleet, vehicle limitations per weight and volume (capacity), vehicle filling mode when goods are sold at the unit level (article), dynamic calculation of the cost depending on the weight of the goods the vehicle is transporting, possibility of multiple and divided travelling for each vehicle, reasonable period of execution of algorithm for normal use in real situations, and other.

**Route cost –** this implies the number of kilometers multiplied by the cost per kilometer of the vehicle driving that route. That is the real route cost. Different penalties are added to that. Sometimes it is allowed for a route not to meet certain constraints, and for each constraint that is not met punishment points (penalties) are added, which increase when the violation of each of the constraints increases.

The real route cost $RRC$ is equal to:

$$RRC = d * c_v \tag{1}$$

$d$ – total distance of route travelled measured in kilometres,

$c_v$ – cost in monetary unit per kilometre for the vehicle driving that route.

This definition implies that only the variable cost of the vehicles is to be considered. In real examples, there are fixed costs as well, which can be described as the cost of each of the vehicles (cumulative amortization, registration, maintenance, tires, vehicle insurance and others) per day. The fixed cost can also include driver cost (one or more of them that are necessary for delivery) for each vehicle separately. If the fixed vehicle cost is taken into consideration as well, then the real route cost changes as follows:

$$RRC = FC_v + d * c_v \tag{2}$$

$d$ – total distance of route travelled measured in kilometres,

$c_v$ – cost in monetary unit per kilometre of vehicle driving that route,

$FC_v$ – fixed cost of the vehicle traveling the given route.

Cost $c_v$ mostly depends on the fuel consumption of the vehicle $v$, while cost $FC_v$ includes vehicle amortization costs, driver costs, and others. Route delays happen when a vehicle is unable to service one or more customers until the end of the time window. If the vehicle arrives before the time window to a customer, it waits for delivery until the start of the time window and then starts delivering.

This delay does not represent route delay. instead, it is used in the context where the stated delays are added, and legally essential drivers' breaks for every vehicle during the workday are obtained. The mentioned delays, if they exist, can be combined with the longest unloading at the customer, and that way necessary drivers' breaks are respected. The penalty for route delay $PD$ is the sum of the delays at all of the customers of that route (presented in minutes) multiplied with the constant $PenaltyDelay$.

$$PD = PenaltyDelay * \sum_k \max(0, d_k - b_k) \tag{3}$$

$d_k$ – the end time of servicing the $k$-th customer,

$b_k$ – the end of the time window of the $k$-th customer.

The penalties for overloading in volume or weight are calculated by taking into account the percentage of the excess, so that 1 [m³] of excess costs bigger vehicles less, considering that in percentages, for bigger vehicles, it affects the problem of exceeding the given constraint less.

The penalty for volume $PV$ is equal to the quotient of the excess in cubic meters and maximum volume, which the vehicle driving the route can carry multiplied with the constant $PenaltyPercentageVolume$.

$$PV = PenaltyPercentageVolume * \max\left(0, \left(\sum_k v_k\right) - mV\right) / mV \tag{4}$$

$v_k$ – the ordered volume of the $k$-th customer,

$mV$ – maximum volume [m³] which the vehicle driving the route can carry.

The weight is penalized analogously as volume, the penalty $PW$ is acquired when the relative excess is multiplied with the constant $PenaltyPercentageWeight$.

$$PW = PenaltyPercentageWeight * \max\left(0, \left(\sum_k t_k\right) - mW\right) / mW \qquad (5)$$

$t_k$ – the ordered weight of the $k$-th customer,

$mW$ – maximum weight [kg] which the vehicle driving the route can carry.

Given that there are certain constraints in which some customers cannot be serviced by certain vehicles (SDVRP), the route cost also includes the penalty for the violation of those conditions. The penalty for customers in the wrong vehicles $PCV$ is equal to:

$$PCV = l * PenaltyCustomersVehicles \qquad (6)$$

$PenaltyCustomersVehicles$ – constant which defines how many penalties are added for one customer in the wrong vehicle,

$l$ – number of customers in the route that cannot be serviced by the vehicle of this route.

The route cost also includes the $PPW$ penalty, which presents the cost increase for vehicles when they are transporting a bigger weight. In real examples of solving the VRP, it has been established that it is necessary to deliver the merchandise to the end customers as soon as possible, given that the consumption of fuel (and other vehicle parts) is significantly larger when the vehicle is transporting a greater amount of merchandise (measured in weight). That is the reason why the parameter $PPW$ was introduced. It strives to primarily service closer customers, especially those who, in percentages, have a greater impact on the total weight of the whole route due to the weight. $PPW$ can be presented with the following equation:

$$PPW = costIncreasing * c_v * \frac{\sum_k d_k \, w_k}{mW} \qquad (7)$$

$costIncreasing$ – constant which presents the coefficient of cost increase when the vehicle is transporting its maximum weight,

$c_v$ – cost of the vehicle $v$ which is travelling the route,

$d_k$ – distance travelled to the $k$-th customer,

$w_k$ – the weight ordered by the $k$-th customer,

$mW$ – maximum weight which the vehicle travelling the route can transport.

In laboratory conditions, this fact is completely disregarded, and the route is primarily considered optimal if customers farther away from the warehouse are serviced first. However, from the aspect of practical application, the given statement is different. If there are customers in a city that is farther away from the warehouse, and a customer (or several) that is very close to the warehouse, but whose ordered weight is, for example, 20% of the total weight of the given route, in laboratory observations the observed vehicle will transport the necessary 20% of route weight from the customer (near the warehouse) to the city and back. Taking into account the fact that the vehicles travelling the routes can be large trucks (with a great transport capacity) and the routes long (several hundred kilometres), then the consumption of fuel in those cases can be larger than in the case where the given customer is serviced among the first (at the start of the route), during departure from the warehouse itself. The given modification decreases the real route cost created using the proposed algorithm. This is one of the contributions of this paper that has not been studied in more detail in other scientific papers.

The total route cost $r$, marked as $T(r)$, is the sum of the real cost and all of the listed penalty costs.

$$T(r) = RRC(r) + PD(r) + PV(r) + PW(r) + PCV(r) + PPW(r) \qquad (8)$$

The cost is calculated this way during the third step of the proposed algorithm (which will be explained in greater details later on), which is conceivably the most important for the optimality of the overall algorithm. During the first step and mid-step of the algorithm delays and route, overloads are not allowed, so those penalties are equal to 0. Also, given that there are no vehicles during the first step of the route, $PKV$ is overlooked. Taking that into account, the cost during the first step is $C_1 = RRC + PPW$, and during the second step $C_2 = RRC + PCV + PPW$.

**Solution cost** – solution cost is the sum of route costs which that solution entails.

$$SolutionCost = \sum_{r=1}^{number\ of\ routes} T(r) \qquad (9)$$

Now that the main terms and their definitions have been stated, the formulation of the problem for the implemented algorithm can be given as well: *for the given set of customers and their orders, vehicles and their characteristics, time constraints related to the servicing of customers, time constraints related to the working hours of each of the vehicles, time constraints related to the warehouse (where all deliveries start and end at one warehouse), constraints related to which vehicle can service which customer, as well as other realistic constraints, it is necessary to find a set of routes and to assign a vehicle to each route, so that the solution cost $SolutionCost$ is minimized. Ideally, the real cost $RRC$ will be minimal, and all the other costs related to penalizing the violation of one or more of the constraints will be equal to zero.*

### 3.1.1 Pre-step – Data initialization and Transformations

Before the transformation of customers' time windows, as is shown in Figure 2, two other types of modifications/transformations are done. The first type relies on the fact that the proposed algorithm has the possibility of defining two additional warehouse parameters:

$prv_0$ – the start of the warehouse's working time (depot's time window), which is the time when the warehouse is available for use. More precisely, the given time is the earliest time the vehicles can leave the warehouse.

$krv_0$ – the end of the warehouse's working time (depot's time window). More precisely, the given time is the last time the vehicles can return to the warehouse after they have serviced the customers from the given route.

Based on the two parameters, the working time of the warehouse, as well as the working time of the vehicles themselves, can be controlled, which presents a complex option (generalization) of this constraint. That is included and explained in a later step of the proposed algorithm. The given parameters are taken into account so that the time windows of each of the customers are corrected through the following iterations:

- The difference $a_i - t_{0i}$ is calculated for each customer, where $a_i$ represents the start of the time window for the customer $i$, and $t_{0i}$ represents the time distance from depot to customer $i$.
- If the difference $a_i - t_{0i} < prv_0$, is given, then the start of the time window of the customer is set to the value $a_i = prv_0 + t_{0i}$. Otherwise the value $a_i$ does not change for customer $i$.
- The sum $t_{i0} + b_i$, is calculated for each customer, where $b_i$ presents the end of the time window for customer $i$, and $t_{i0}$ presents the time distance from customer $i$ to the depot.
- If the sum $t_{i0} + b_i > krv_0$ is given, then the end of the time window for customer $i$ is set to the value $b_i = krv_0 - t_{i0}$. Otherwise the value $b_i$ does not change for customer $i$.

The second transformation is based on the fact how the proposed algorithm is supposed to perform multiple deliveries to the customers in the situations when customer cannot be serviced by a valid vehicle in one delivery while not simultaneously violating SDVRP constraints. In order to meet those constraints, the algorithm performs the following iterations during this pre-step of data preparation:

- For each customer, a set of available vehicles that can service them are checked, and then they are stored into a list of vehicles available to the customer
- From the list of vehicles available to each customer a comparison is made between the volume and transport capacity of the vehicle and the volume and weight of the customer's order
- If the list of available vehicles to the customer contains vehicles, whose volume and transport capacity are greater or equal to those of the volume and weight ordered by the customer, the iteration for that given customer is stopped, and the next customer is reviewed
- If the list of available vehicles to the customer does not contain vehicles whose volume and transport capacity are greater or equal the volume and weight of the customer's order, several sub-iterations are done within this iteration:
    - For each allowed vehicle for multiple deliveries, the profitability of the allowed vehicle for the given customer is calculated the following way:

$$profitabilityAllowedVehicle = \left\lfloor max\left(\frac{v_k}{mV}, \frac{w_k}{mW}\right) \right\rfloor * c_v \tag{10}$$

– The vehicle which is most profitable is chosen, and the number of travels of the given vehicles to the customer is equal to $\left\lfloor max\left(\frac{v_k}{mV}, \frac{w_k}{mW}\right) \right\rfloor$, and those routes are created in advance. The remaining part of the order $\left\{ max\left(\frac{v_k}{mV}, \frac{w_k}{mW}\right) \right\}$ of the customer remains a candidate for routing in the following steps of the proposed algorithm. The start of the working time of the most profitable vehicle $prv_v$ is set to the value:

$$prv_v^{'} = prv_v + \left\lfloor max\left(\frac{v_k}{mV}, \frac{w_k}{mW}\right) \right\rfloor * \left( t_{i0} + \frac{s_i}{max\left(\frac{v_k}{mV}, \frac{w_k}{mW}\right)} + t_{0i} \right) \tag{11}$$

– The unloading time for the customer $s_i^{'}$ which will be used in the following steps of the proposed algorithm is equal to:

$$s_i^{'} = \left\{ \frac{s_i}{max\left(\frac{v_k}{mV}, \frac{w_k}{mW}\right)} \right\} \tag{12}$$

– The time window of the customer for use in later steps of the algorithm is equal to:

$$a_i^{'} = max(a_i, prv_v^{'})$$
$$b_i^{'} = min(b_i, krv_v) \tag{13}$$

– When all customers are checked, the iterations are stopped

Besides the listed transformations, other data preparations are done before the start of the algorithm's execution. The listed preparations of the algorithm's input data will be explained later in the following sections of this work.

Transformation maps the current problem into a different problem which can be proven to be the same as the first but is more simple to realize. This pre-step only change the algorithm's input data related to time. After the transformation is done, the unloading time for each customer is 0, and the time distances between the customers have increased.

The implemented transformation is published in the work of the authors Liu and Shen [36], and it can be implemented if the distances meet the triangle inequality theorem, which is true for real-world problems of transport route optimization. Time distances for every two customers increase for half the unloading duration of those customers (intuitively, half the unloading time for each customer is assigned to the time distance to that customer, and the other half is assigned to the time distance from that customer). Time distances for each customer from the depot increase for half the unloading time of that customer. After that, time windows change as well, so that half the unloading time for that customer is added to the start of the time windows for each customer and the end of the time window decreases by half the unloading time for that customer. Meaning:

$$t_{ij}^{'} = t_{ij} + \frac{s_i}{2} + \frac{s_j}{2}, t_{i0}^{'} = t_{i0} + \frac{s_i}{2}, t_{0i}^{'} = t_{0i} + \frac{s_i}{2}, \left(a_i^{'}, b_i^{'}\right) = \left(a_i + \frac{s_i}{2}, b_i - \frac{s_i}{2}\right) \tag{14}$$

where $s_i, t_{i0}, t_{0i}, a_i$ and $b_i$ are unloading times of the customer, time distance from depot to customer $i$, start of the time window and end of the time window of the customer $i$, in that order. $t_{ij}$ is the time distance of the customers $i$ and $j$, and $t_{ij}^{'}, t_{0i}^{'}, t_{i0}^{'}$ and $\left(a_i^{'}, b_i^{'}\right)$ are the new values of od $t_{ij}, t_{i0}, t_{0i}$ and $(a_i, b_i)$ after the transformation, in that order.

After these changes are implemented, the unloading times for each of the customers $s_i$ are equal to 0. The newly acquired time distances between the customers are stored into a new matrix $newDistancesT$, and all of the data related to the customers, with an exception of the unloading times, are changed in the list of customers. The unloading times only change after the routes are made, locally at the customers of the routes, so that it is possibly to return the data to an initial state using inverse transformation (so that the value of unloading duration is not lost).

The implemented transformation is not completely the same as in the mentioned research. Instead, it is adjusted to be accurate in this case. The difference is that the work assumes that the delivery starts in the time window, and does not necessarily have to end in the time window, which is not accurate in the algorithm presented in this work. Also, in the mentioned work, the working time of the warehouse is changed in this step as well and considering that the given modification was done in the pre-step of the proposed algorithm, it is not done again in the transformation step.

The problem must stay the same, and after the algorithm is executed, and before the final print out of the results, an inverse transformation is done as well, so that all data is returned to the initial state understandable to the end-users. The inverse transformation is easily derived from the transformation (where quantity was added, it is now reduced for that amount, and where it was reduced by a certain amount, it is now added).

### 3.1.2    First step – Generating of initial routes using Clarke and Wright savings algorithm

Clarke-Wright heuristic (or Clark and Wright savings algorithm) starts with one route created for each customer. Every route goes from the warehouse to the customer and back to the warehouse. After that, savings are calculated for every two customers with the following formula: $u_{vivj} = c_{viv0} + c_{v0vj} - c_{vivj}$. It shows savings measured in distance travelled, when two routes, whose end customers are $v_i$ and $v_j$, are merged. This means that vehicle after servicing customer $v_i$, instead of going to the warehouse, it now goes directly to customer $v_j$ and goes to the warehouse after. That way savings are acquired. It is often assumed that the distances of the VRP satisfy the triangle inequality theorem, $c_{vivj} \leq c_{vivk} + c_{vkvj}$, for all indexes $0 \leq i$; $j$; $k \leq n$. In those cases, all of the savings are clearly non-negative.

After all savings are calculated, they are sorted in descending order. After that, the following is done in the series of iterations (Pseudocode 2):

- The largest savings that have not been observed are taken.
- It is checked whether the observed customers are in different routes and whether the end customers (first or last) are in their routes. If at least one of these is not satisfied, the iteration is done.
- If the weight of the route exceeds the capacity $Q$, the iteration is done.
- Two routes are merged by merging the observed customers.

After these iterations are finished, the current set of routes is the result of the Clarke and Wright savings algorithm. This algorithm was originally written for the CVRP but has been later implanted on other variants of the VRP. It is mostly modified in a way where other conditions, which need to be met for this variant of the VRP, are checked while merging routes. For example, with the VRPTW, besides checking other conditions (for route capacity), it would be sufficient to also check whether the new route is feasible concerning time constraints.

As it was said, the initial solution is created in the first step. From the main version of the Clarke and Wright algorithm, other versions have been derived over time, some of which give better results. One of those was formed in 1999, in the work of Liu and Shen [36], and that version is implemented in the proposed algorithm. The main difference, from the main version, is that it not only observes the possibility of merging two routes but also observes the possibility of inserting a route between two customers on a different route. Besides that, the insertion of an inverse route between two customers of a different route (a reversed route is acquired by reversing the order of servicing customers), is observed as well. Given that one route can be inserted between two customers of a different route, savings cannot be calculated in advance, as is the case with the general algorithm. Therefore the algorithm is slightly different as well. The initial solution is acquired the same way, but the iterations are performed differently. In each iteration, two routes are merged. Of all the merging possibilities, the one with the highest savings is selected and executed.

During the selection of the routes, each pair of routes that can be merged is observed, taking into account that the new route needs to meet these conditions: weight, volume and time of arrival at the customer. The first step does not allow for a route to have a weight larger than the constant $BIGGEST\_CAPACITY\_WEIGHT$, which is set to the value of the largest vehicle's weight capacity. Also, it is not allowed for a route to have a volume greater than the constant $BIGGEST\_CAPACITY\_VOLUME$, which is set to the value of the largest vehicle's volume capacity. Each overrun is calculated assuming that the largest vehicle will be driving each route. On the other hand, in the first step, the proposed route would have to depend on the available fleet of vehicles. To take into account the vehicles, route cost, which up to now only depended on the travelled distance, now depends on the fleet of vehicles. The route cost is equal to the distance travelled multiplied by the variable cost of the cheapest vehicle that can drive that route (by capacity).

---

**Pseudocode 2: Generating of initial routes using Clarke and Wright savings algorithm**

```
while(num_of_iterations--){
  double maxSavings = - CONSTANT; int maxI,maxJ,maxK; bool inverted;
  for(int i = 0; i < routes.size()-1; i++){
    Route *r2 = routes[i]->GetInvertedRoute();
    for(int j = i+1; j < routes.size(); j++){
      for(int k = -1; k < 0 || k < (routes[i]->customers.size()); k++){
          if(routes[i]->IsItPossibleToInsertRouteToThePlace(routes[j],k) &&
            maxSavings < routes[i]->ModifiedSavings(routes[j],k) -
                  TotalDifferencConstraintsPerVehicles (routes[i], routes[j], vehicles.size())){
              maxI = i; maxJ = j; maxK = k; inverted = false;
              maxSavings = routes[i]->ModifiedSavings(routes[j],k);
          }
      }
      for(int k = -1; k < 0 || k < (routes[j]->customers.size()); k++){
          if(routes[j]->IsItPossibleToInsertRouteToThePlace (routes[i],k) &&
            maxSavings < routes[j]->ModifiedSavings(routes[i],k) -
                  TotalDifferencConstraintsPerVehicles (routes[i], routes[j], vehicles.size())){
              maxI = j; maxJ = i; maxK = k; inverted = false;
              maxSavings = routes[j]->ModifiedSavings(routes[i],k);
          }
      }
      Route *r = routes[j]->GetInvertedRoute();
      for(int k = -1; k < 0 || k < (routes[i]->customers.size()); k++){
          if(rute[i]->IsItPossibleToInsertRouteToThePlace(r,k) &&
            maxSavings < routes[i]->ModifiedSavings(r,k) -
                  TotalDifferencConstraintsPerVehicles (routes[i], routes[j], vehicles.size())){
              maxI = i; maxJ = j; maxK = k; inverted = true;
              maxSavings = routes[i]->ModifiedSavings(r,k);
          }
      }
      for(int k = -1; k < 0 || k < (routes[j]->customers.size()); k++){
          if(routes[j]-> IsItPossibleToInsertRouteToThePlace(r2,k) &&
            maxSavings < routes[j]->ModifiedSavings(r2,k) -
                  TotalDifferencConstraintsPerVehicles (routes[i], routes[j], vehicles.size())){
              maxI = j; maxJ = i; maxK = k; inverted = true;
              maxSavings = routes[j]->ModifiedSavings(r2,k);
          }
      }
    }
  }
  if(maxSavings < 0.01) break;
  if(!inverted)
      routes[maxI]->InsertRouteToThePlace(routes[maxJ],maxK);
  else
      routes[maxI]-> InsertRouteToThePlace(routes [maxJ]->GetInvertedRoute(),maxK);
  routes.erase(routes.begin()+maxJ);
}
```

When it comes to the SDVRP problem, a new term (constant) was introduced in the first step, which shows the similarity between routes and which is affected by constraints related to which vehicle cannot service which customer. During implementation and testing of the algorithm, it can be noticed that customers, who cannot be serviced by a certain vehicle are in several different routes, so that vehicle cannot travel any of those routes. It happened regularly that a set of routes cannot be travelled by an available fleet of vehicles. That's why the algorithm allows the option to reduce the savings by a number (constant) *routesSimilarity* which shows how similar routes are in terms of constraints, and increases the chance that two routes with similar constraints are merged. Taking into account the number *routesSimilarity*, the algorithm tends to put all of the customers that cannot be serviced by a certain vehicle into the same route, so that only one route cannot be travelled by that vehicle. The number *routesSimilarity* is obtained, so that for each pair of customers, where one customer is in one route and the other is in the second route, 1 is added to each of the differences in constraints of those two customers. This is one of the secondary contributions of this paper.

Each customer has information for each vehicle, regarding whether it can be served by them. Each vehicle, which can serve only one of two customers, represents the difference. After the number of differences is calculated, that number is divided by the sum of the number of customers in those two routes (which normalizes this penalty) and is multiplied by the constant, which often had the value 3. After the highest savings amount is determined, if it's possible to obtain savings by merging (there is a chance that no two routes can be merged), the routes are merged (if the highest savings amount was obtained when a reversed route was inserted into another, the reversed one is inserted). That way, after each iteration, two routes are merged, and the number of routes decreases by 1. If it is not possible to merge any two routes to obtain savings, the algorithm terminates.

### 3.1.3   First step – Assignment of vehicles to the routes

If it is not possible to merge two routes to obtain savings (the savings will be negative), then the step terminates. The current set of routes is the solution. Because the routes do not have their vehicles yet, in the following step vehicles are assigned to the routes using a brute-force approach.

All possible permutations of vehicle assignments are observed, and the one with the minimal cost is chosen (Pseudocode 3). The number of permutations increases rapidly, and this method is used when the number of available vehicles in the fleet is less than 10-15 (the number of permutations is reasonable, based on experimental measurements). For fleets with a vehicle count greater than 10-15, vehicle assignment is done using the greedy algorithm where the number of permutations quickly increases. First, the vehicle with the lowest cost is assigned to the longest route, from the vehicles with a capacity large enough to travel that route. In following iterations, a vehicle with the smallest cost is assigned to a route. If the number of routes is greater than the number of vehicles, a necessary amount of fictitious vehicles is included. A fictitious vehicle has cost larger than the other vehicles. Its cost is a constant *ADDITIONAL_VEHICLE_COST*, and its capacity is represented using two constants: *BIGGEST_CAPACITY_WEIGHT* and *BIGGEST_CAPACITY_VOLUME*.

```
Pseudocode 3: Assignment of vehicles to the routes

double costBestAssignment = CONST_BIG_NUMBER;
// additionalVehicle is added if there aro no enough vehicles in the fleet
for(int i = 0; i < routes.size(); i++){
  currentAssignment[i] = i;
}
do{
  vector<Route*> currentAssignment;
  for(int i = 0; i < routes.size(); i++){
    if(currentAssignment[i] < vehicles.size())
      currentSolution.push_back(new Route(routes[i]-> customers, vehicles[currentAssignment[i]],true));
    else
      currentSolution.push_back(new Route(routes[i]->customers,additionalVehicle,true));
  }
  double currentCost = SolutionCost(currentSolution);
  if(currentCost < bestCost){
      for(int i = 0; i < routes.size(); i++){
```

```
      bestAssignment[i] = currentAssignment[i];
    }
    costBestAssignment = currentCost;
  }
  for(int i = 0; i < routes.size(); i++){
    for(int j = 0; j < routes[i]->customers.size(); j++){
      delete currentSolution[i]->customers[j];
    }
    delete currentSolution[i];
  }
} while(next_permutation(currentAssignment, currentAssignment+routes.size()));
for(int i = 0; i < routes.size(); i++){
  if(bestAssignment[i] < vehicles.size())
    routes[i] = new Route(routes[i]->customers, vehicles[bestAssignment[i]], true);
  else
    routes[i] = new Route(routes[i]->customers,additionalVehicle,true);
}
```

The following step of the algorithm needs to remove this vehicle (because if it does not, then the algorithm does not return a valid solution). In practice, a great cost of this vehicle forces the second step to remove it, and if it does not succeed in removing it, then it is intuitive and impossible to find the solution with a real fleet of vehicles.

### 3.1.4   Second step (intermediate step) – Route elimination

When the solution cost is observed by taking into account the number of routes, it can be noticed that the lower the number of routes is, the lesser the solution cost is as well. Even though that does not always have to be the case, the heuristic algorithms for the VRPTW always assume that the solution with the smaller number of routes (or used vehicles) is more optimal than the solution with a greater number of obtained routes (or used vehicles), and the cost is only observed when two solutions have the same number of routes (or used vehicles in the routing plan). One of the possible reasons, why this is the case, is because every route has its cost, which is independent of variable cost (fixed cost), and so it is better to have a smaller number of routes. In the case of the HVRPTW problem, which is most commonly found in real-world examples; this is not the case, considering that different vehicles travel the routes, the number of routes do not mean much.

After the first step of the algorithm has returned a set of routes, this step of the algorithm attempts to decrease the number of routes. The solution with fewer routes is accepted only if the cost is less or equal to the initial cost. The idea of route elimination was taken from the work of Braysy et al [37]. Route elimination is done using a constant number of iterations, and in each of the iterations the following is done (Pseudocode 4):

  – Chooses one permutation using the current set of routes.
  – Using the first route, permutation respectively tries to put each customer into another route.
  – If it succeeds in putting all of the customers in other routes, and the cost of the newly obtained solution is less than the cost of the solution before, the route is removed from the permutation and the previous step is done. If it does not succeed in doing that and obtaining lesser cost, then all of the customers are put in the initial route and the next iteration is done.

---
**Pseudocode 4: Route elimination**

```
while(num_of_iterations --){
  vector<int> randPermutation = RandomPermutation(currentSolution.size());
  for(int i = 0; i < randPermutation.size(); i++){
    vector<Route*> routeCopy = currentSolution;
    for(int j = 0; j < routeCopy.size(); j++){
      routeCopy[j] = new Route(currentSolution[j]->customers, currentSolution[j]->routeVehicle,true);
    }
```

```
     double totalConsumption = 0;
   bool possible = true;
   for(int k = 0; k < routeCopy[randPermutation[i]]->customers.size(); k++){
      double minConsumption = CONSTANT;
      int route_number = -1, place = -1;
      for(int u = 0; u < routeCopy.size(); u++){
         if(u != randPermutation[i]){
            for(int v = -1; v == -1 || v < routeCopy[u]->customers.size(); v++){
               if(routeCopy[u]->IsItpossibleToAddCustomer(routeCopy[randPermutation[i]]->customers[k],v)
                  && minConsumption > routeCopy[u]-> AdditionalConsumptionwithCustomerAtPlace
                  (routeCopy[randPermutation[i]]-> customers[k],v)){
                     minConsumption = routeCopy[u]-> AdditionalConsumptionwithCustomerAtPlace
                        (routeCopy[randPermutation[i]]->customers[k],v);
                     route_number = u;
                     place = v;
               }
            }
         }
      }
      if(route_number == -1){
         possible = false;
         break;
      }
      routeCopy[route_number]->
         InsertRouteToThePlace(new Route(routeCopy[randPermutation[i]]->customers[k]),place);
      totalConsumption += minConsumption;
   }
   double savings = currentSolution[randPermutation[i]]->RouteCost();
   if(possible && (savings - totalConsumption > 0)){
      routeCopy.erase(routeCopy.begin()+randPermutation[i]);
      currentSolution = routeCopy;
      break;
   }
 }
}
```

Each permutation has a chance to be chosen with equal probability, and the way of choosing permutations is explained in the book of Skiena [38]. When a customer is put in a different place, all possible places where that customer can be put are checked, and the one with the smallest cost is selected. If, in the end, the total sum of costs of rearranging customers is less or equal to the cost of the route that is removed, the removal of the route is approved, and the next route in the permutation is observed. That route is now the first route of the permutation.

### 3.1.5   Third step – Tabu search

Tabu search, created by Fred W. Glover in 1986 [39] and formalized in 1989 [40]-[41], is a metaheuristic search method employed as a local search method used for mathematical optimization. Local search is a method of solving problems of combinatorial optimization which start from one solution, and in a series of iterations, in each iteration, it goes from one solution to a solution in the neighbourhood of that solution. The neighbourhood of the solution presents solutions which are, in some way, similar to that solution and is most often obtained like other solutions which can be obtained from the current solution through some form of modification.

Naive local search always chooses the best solution from the neighbourhood of the solution, but doing so can cause repetitions of the solutions (one of the possibilities is that the search finds two solutions that are the best solutions in that neighbourhood). Tabu search solves that problem by forbidding some solutions from the neighbourhood, which prevents the repetition.

Those solutions can be forbidden by remembering the list of forbidden solutions, or, as is the case with this algorithm, by remembering the component of the solution that is forbidden. The proposed algorithm forbids a customer to be in a route a certain number of times after ending up in that route. That way, all solutions which have the customer in that route are forbidden. That way, it is forbidden for solutions to start repeating themselves, and the algorithm tends not to repeat even similar solutions so that it can enter a different area of possibility when it comes to solutions.

After the previous part is finished, the third step starts, which is the Tabu search with the current solution as the initial solution. The main idea was taken from the work [42], and the algorithm was continuously improved.

The Tabu search procedure, which has a fixed number of iterations, is started. The current solution changes in each iteration. The whole time, the best solution (to the current iteration) is memorized, and in each iteration, it is checked whether the current solution is better than the previous one, and if it is, then the current solution becomes the best solution. At the end of the iterations, the best solution becomes optimal (Pseudocode 5).

For the neighbourhood of solutions, two operators which modify the solution are used: *RELOCATE* and *SWAP*. The procedure will be briefly explained in the following text.

The operator *RELOCATE* removes a customer from their route and relocates them to a different route. All possible combinations are observed, and the one with the highest savings is selected. The algorithm goes through multiple iterations, one iteration for each customer. In every iteration, the following is done for customer $k$:

- The savings $U$ are calculated for the cost, in the case where the customer $k$ would be relocated from their route.
- For each route except the one $k$ is in, the following is done:
  - The place where the customer $k$ would be relocated to, which gives the least cost $T$, is found.
  - If $U - T$ is the biggest so far, this relocation and the savings are memorized as the highest.

After all of the iterations are done, the relocation which gave the highest savings is performed, provided those savings are higher than 0. Solutions which have overruns in terms of weight and volume, as well as delays, are permitted, but additional penalty points exist for not meeting certain constraints, as it is described in the definition of the solution cost, at the beginning of this section. For each customer, the number of places where the customer can be relocated to, is approximately equal to the number of customers, considering that the customer can be relocated behind every customer, except the ones from their route and it can also be relocated to the start of each route except for its current route (so the number of possible places is $n - t + k - 1$, where $n$ is the number of customers, $k$ is the number of routes, and $t$ is the number of customers in the current route of the customer). If there are $n$ customers, every one of the removed customers can be relocated to approximately $n$ places. Therefore, every solution has $O(n^2)$ solutions in its neighbourhood when the *RELOCATE* operator is used, and that also determines the time complexity of the operator.

The SWAP operator exchanges two customers from different routes. Similarly to the previous operator, every two customers, who are not from the same route, are observed, as well as the savings, which would be obtained should the relocation be performed. In the end, two customers who give the highest savings are chosen. If those savings are greater than 0, the exchange is performed. The number of possible customer relocations depends on the number of routes and the number of customers per route. Considering that the number of routes is always at least 3, and the number of customers per route is mostly approximately equal for every route, each customer can be relocated with at least $n/2$ of the rest of the customers (because their route does not have more than $n/2$ customers). So there are at least $(n * n/2)/2 = n^2/4$ possible relocations (every customer relocates with at least $n/2$ others, and is divided by 2 because each relocation is counted 2 times). Therefore, for each solution, there are $O(n^2)$ other solutions that can be obtained by relocating two customers.

Following the presented methods and the operators, in each iteration, the proposed algorithm observes how *RELOCATE* and *SWAP* can be changed and improves the solution. The method, which gives the best solution and is not forbidden in terms of the set constraints, is selected.

The algorithm has characteristics of Tabu search in a way that it forbids the customer to enter the same route twice in a short amount of iterations. That is achieved by keeping a list (matrix) in which rows represent customers, and columns represent routes (called *tabuValuesMatrix* and which store the value of how many subsequent iterations a certain customer is forbidden from entering a certain route, while combinations which are not allowed during calculation of cost are not taken into account.

---

**Pseudocode 5: Tabu search – simplified view**

---

```
Initialization tabu_list;
Initialization x; // initial solution of this step
Set_of_the_solutions X = ∅;
while (stopping_criteria != true){
    x={y|y not in tabu_list or addition_criteria(y) == true};
    If(X == ∅) exit_from_the_loop;
    x = choose_the_best_solution(X);
    update_tabu_list();
    update_addition_criteria();
}
```

---

Every time a customer enters a route, with the operation RELOCATE, the assigned value to that customer and that route becomes equal to the constant TABU, which represents the number of following iterations in which that customer will not be able to enter that route. Every time the SWAP operation happens, and two customers enter new routes, the values in the matrix *tabuVrijednosti,* which represent those two customers and their new routes are set to value $TABU/2$. The SWAP operation is not executed only when both customers are forbidden from entering routes, which they would otherwise enter. After selecting the modified solution with the highest savings, the operator is executed. Then, when the operator RELOCATE is applied, the vehicles are assigned to their routes again, using the greedy algorithm. The routes are sorted by length first, and vehicles are sorted by travelled kilometres. Afterwards, the cheapest vehicle from the fleet, which can travel that route in terms of capacity (weight and volume) is assigned to the longest route, and that vehicle becomes unavailable. The process is then continued, and the longest route gets assigned the cheapest available vehicle in terms of capacity which can travel, and the process continues with the third-longest and the rest of the routes until all iterations are completed. Before the assignment is done, solution cost is memorized, and if the solution cost is greater after the assignments, the routes are assigned to the vehicles like they were before, and the new assignment is cancelled.

Each 1000th iteration, all possibilities when it comes to the assignment of vehicles to routes are checked, and the best one is chosen. Given that checking all possibilities is a demanding operation, it is not possible to try all the possibilities from every iteration, and that is why vehicles are usually assigned using the *greedy* algorithm described earlier. This is one of the secondary contributions of this paper.

Also, the proposed algorithm tries to research a greater area of solutions and strives to find more diverse solutions. It does that by calculating how many times the pair (customer, route) appeared in the solution (how many times the route had that customer). That is stored in the table *numberOfInclusions,* and if the cost in the transition is positive, the number of times the customer was in the route multiplied by the appropriate constants is added to the cost. Intuitively, this means that only when the current solution does not have a better solution than itself in its neighbourhood, during the calculation of the transition cost, the solutions which appeared less up until that moment are preferred, so that the algorithm has a more diverse search. The idea given in [42] is that for each pair (customer $k$, route $r$) the best solution in which that pair appears in (in which customer $k$ is in route $r$) is memorized as well. When it happens that a solution $R$ cannot be the next solution because customer $k$ cannot enter route $r$ because of *tabuValuesMatrix,* two options are considered:

- The solution cost $R$ is not observed at all, which is the simplified form.
- The solution cost $R$ is calculated, and if that cost is less than the minimal cost so far, in which customer $k$ and route $r$ participate, then the solution $R$ is chosen.

The second form presents a modification of the standard Tabu search.

### 3.1.6 Fourth step (post-step) – Improvements within each route

The fourth step of the whole modular algorithm additionally strives to optimize each route separately. Given that the previous step (Tabu search) relocates the customers between different routes, situations often occur where routes which are the result of the third step can be additionally improved, and in a way where customers are relocated from one location to another, within the same route. Besides that, considering that the algorithm is conceived to solve practical problems, where the route cost directly depends on the weight of the merchandise the vehicles are transporting and delivering to each customer, with permutations of the customer within each route, this fact can be fixed.

This statement can be especially noticed when a part of a certain route is located in one city, and the other part in another city, which is farther away from the warehouse.

The fourth step could be described in the following way, in terms of iterations (Pseudocode 6):

- Take one customer from the route and try to relocate him to every place in the same (his) route.
- Calculate the route cost for every customer relocation.
- If the newly calculated route cost is less than the route cost before the relocation of the customer, the route becomes optimal. Otherwise, the optimal route is the initial route in this step, and the customer is returned to the starting position in the route.

**Pseudocode 6: Improvements within each route**

```
for(int i = 0; i < num_of_iterations; i++){
  int customer = -1;
  int place = -1;
  double theLowestCost = RouteCost();
  for(int j = 0; j < customers.size(); j++){
    vector<Customer*> newRoute;
    for(int r = 0; r < j; r++){
      newRoute.push_back(customers[r]);
    }
    for(int r = j+1; r < customers.size(); r++){
      newRoute.push_back(customers[r]);
    }
    for(int k = 0; k <= newRoute.size(); k++){
      newRoute.insert(newRoute.begin()+k,customers[j]);
      Route *r = new Route(newRoute,routeVehicle,true);
      double routeCost = r->RouteCost();
      if(routeCost < theLowestCost)
          customer = j; place = k; theLowestCost = routeCost;
      for(int u = 0; u < r->customers.size(); u++){
        delete r->customers[u];
      }
      delete r;
      newRoute.erase(newRoute.begin()+k);
    }
  }
  if(customer == -1){
    break;
  }
  Customer *k = customers[customer];
  ExcludeCustomer(customer);
  InsertRouteToThePlace(new Route(k),place-1);
}
```

Iterations are finished when there are no possible customer relocations (of one customer) within the same route that decrease the whole route cost. This, of course, implies that all the set constraints are met. Considering that the listed operations are not complex nor demanding in terms of time since they only consider the route cost calculation after permutating the customer, the possibility for additional improvement is noticed, for example, if the same procedure is executed for two or three consecutive customers within the same route. In that case, there is an attempt to relocate two consecutive customers to different sequences, and if the cost is smaller than before, that route becomes optimal, otherwise, after relocating customers one by one (the previously described procedure), the initial route is chosen. There is a small probability that improvements can be achieved if more than three consecutive customers are permutated. For each of the obtained routes, the algorithm terminates at those iterations that relocate three consecutive customers. This way of improvement within each route presents one of the contributions of this paper.

## 3.2   Data-driven approach for adjusting the control parameters of the proposed algorithm

The previous section presents a modular algorithm, which is able to solve real-world VRP with a great number of various realistic constraints. The proposed algorithm consists of control parameters (constants) which are listed in the following text.

***BIGGEST_CAPACITY_WEIGHT*** and ***BIGGEST_CAPACITY_VOLUME*** – During the first step, vehicles are not assigned to routes. However, it is still not permitted for routes to have a total weight that is greater than the weight of the largest vehicle in the fleet, and analogously, routes cannot have a total volume greater than the volume of the largest vehicle travelling that route. These two constants represent those two values which no route can exceed.

***ADDITIONAL_VEHICLE_COST*** – This constant represents the cost per kilometre of the fictitious vehicle, which is added to the solution when a route cannot be travelled by any real vehicles. That can happen when the number of routes is greater than the number of vehicles or when a route is overloaded in terms of volume or weight. The most common value of this parameter is equal to 2.

***NUMBER_OF_ITERATIONS*** – This constant represents the number of iterations of Tabu search. It is currently set through the input parameter of the algorithm, and its most common value is 25000.

***ToleranceWeight*** and ***ToleranceVolume*** – During the *greedy* assignment of vehicles, after each iteration of Tabu search, it is allowed for the vehicle to be overloaded in terms of weight and volume by the values *ToleranceWeight* and *ToleranceVolume*. The most commonly set value for the parameter *ToleranceWeight* is 50 [kg], and for *ToleranceVolume* it is 0.1 [m$^3$].

***TABU*** – Constant related to the execution of the Tabu Search algorithm. The most commonly set value of this constant is 30.

***PenaltyDelay*** – Penalization constant of a one minute delay for the customer. The most commonly set value of this parameter is 0.5 [CurrencyUnit/min].

***Lambda*** – Constant which enables a better search of the solution area. The most commonly set value is 0.001.

***PenaltyCustomersVehicles*** – Constant which penalizes the customer in the vehicle which cannot service them (SDVRP). This parameter is very important meeting the constraints and its most common value is 400.

***CostIncreasing*** – Constant which is used in the cost increase due to the weight the vehicle is carrying. It represents the coefficient of cost increase when the vehicle is transporting the maximum weight. The set value is 0.2.

***PenaltyPercentageVolume*** – Constant which penalizes the overload of a vehicle in volume. It represents the cost increase when the volume is overloaded by 100%. The most commonly set value is 400.

***PenaltyPercentageWeight*** – Constant which penalizes the overload of a vehicle in weight. It represents the cost increase when the weight is overloaded by 100%. The most commonly set value is 400.

Which routes will be obtained as a result, how much the total cost will be, and whether all the set constraints will be met, all depend on the values of the given parameters. The primary goal with real-world VRP is to not violate any of the set constraints. Following that, some of the listed control parameters will be set in a way that is described in the following text, using several prediction methods and algorithms, which use historical data. In short, the whole idea of parameter adjustment can be presented as on the Figure 2.

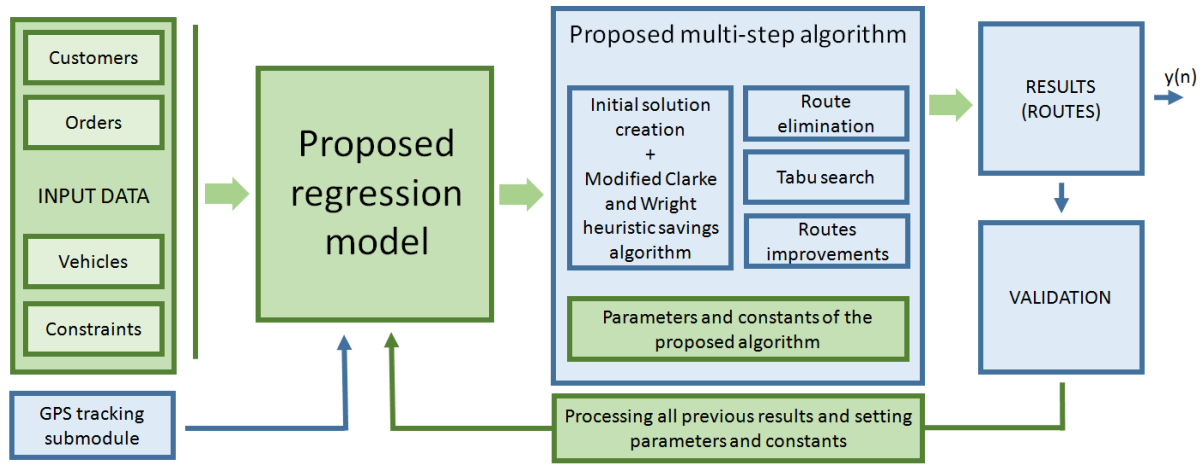

Figure 2: Control parameter adjustment approach using historical data

Based on all the parameters that can influence the final result, several fundamental ones, which are memorized for every route, have been isolated, during testing and in production use of several distribution companies, which deal with product distribution from their warehouses to the delivery locations (markets, shops, supermarkets, and others). That way, a knowledge base, which is updated every day, is created, and the primary goal of this step of the system is to adjust the control parameters and constants, which are an integral part of the system, according to that historical data. With that, the whole approach gets an adaptive character, because the algorithm is improved in time, in a way the parameters, which enable obtaining the minimal cost of the transport route, are adjusted automatically while meeting all of the set constraints at the same time.

Attributes which are isolated as those that affect the obtained routes are:

- number of customers
- number of available vehicles
- number of different available types of vehicles
- number of different cities
- the total number of constraints related to which customer cannot be serviced by which vehicle
- the total number of articles ordered [items]
- the total volume of all articles ordered [m$^3$]
- the total weight of all articles ordered [kg]
- the total duration of time windows of all the customers [min]
- whether all set constraints are met ($1 -$ yes, $0 -$ no)

The goal attributes which affect the obtained routes and total cost, and which represent the control parameters of the implemented algorithm are:

- *ToleranceWeight*
- *ToleranceVolume*
- *PenaltyDelay*
- *PenaltyCustomersVehicles*
- *CostIncreasing*
- *PenaltyPercentageVolume*
- *PenaltyPercentageWeight*

Each of these parameters is separately set. First, the pre-processing of the data was done, and only the data, which meets all the constraints (value 1 in the given column), was selected from history. After, the redundant attributes were removed. Using the *Attribute Importance* option (step), the importance of the input attributes, for every goal attribute, was determined. For attribute importance, *Minimum Descriptor Length* (MDL) algorithm was used. Before that, normalization of the attributes was done so that the attribute values were rounded up to one decimal. After pre-processing and preparation of the input data, the regression model is created for determining the goal attributes. The model is identical for every goal attribute. The proposed model for one parameter is shown on Figure 3.

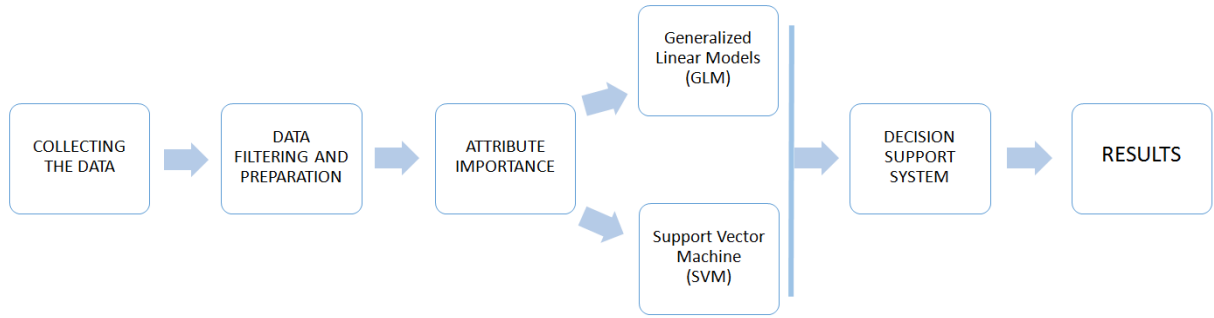

Figure 3: Proposed regression model for one parameter

Two algorithms were used for the regression:

‒ Generalized Linear Models (GLM)
‒ Support Vector Machine (SVM)

These two regression models were selected for several reasons. The advantage of SVM over other methods is that it provides better predictions with unseen test data, provides simple optimal solutions for the problem in training, and there are fewer parameters for optimization in comparison with other methods. The execution speed is not so crucial for the problem that it is used for, and so the disadvantages of the SVM regression method can be ignored. The GLM algorithm for regression was chosen because it represents a generalization of linear regression and is often used in cases when the output variables do not have a normal distribution. Given that the input data are associated with linear dependence, the choice of the GLM regression algorithm was a logical one. After that, the Decision Support System was made, which, based on the obtained results for both of the listed algorithms for each attribute, chooses the one that gives better results by the obtained indicator *Predictive Confidence* [%], and the same is used for the goal attribute during new routing.

The following section lists the obtained results for this part of the control parameter adjustment of the algorithm, the obtained results of the whole proposed VRP modular algorithm on standard data sets, as well as actual data sets, in great detail.

# 4 Discussion and comparison of results

This section will be divided into four subsections:

‒ (4.1) analysis of the obtained results on standardized input data,
‒ (4.2) analysis of the results on real-world benchmark data,
‒ (4.3) analysis of the control parameter adjustment results of the proposed algorithm (from Section 3.2),
‒ (4.4) practical significance of the proposed approach.

## 4.1 Analysis of the obtained results on standardized benchmark data

For VRP with time windows, some data instances have become standard over time and using such data every VRP algorithm is tested and validated. First, in 1987, in [43] Solomon published a set of VRPTW instances, which contain 25, 50, and 100 customers. For a long time, these instances proved to be a challenge for scientists, and there were not any instances with more customers. For the last 20 years, as algorithms for more delivery sites have been developed, there has been a need for instances with more customers. In 2005, Gehring and Homberger [44] created a new set of instances consisting of 200, 400, 600, 800, and 1000 customers each. They generated their instances in the same way Solomon created his. All of the instances are found on the web-page [45]. Optimal solutions are updated regularly on the mentioned page. Those solutions were used in the first method of testing the proposed algorithm of this work.

Solomons' instances consist of $n$ vehicles in the fleet, where $n$ is always equal to one-quarter of the number of customers in the instance, capacity of the vehicles $Q$, information about the depot and customers. Every customer $i$ has his coordinates $x_i$ and $y_i$, ordered weight $q_i$, start and end of the time window $a_i$ and $b_i$, and the unloading time $s_i$. Coordinates $x_0$ and $y_0$ are given for the depot and time window $a_0$ and $b_0$. Distances between customers and the depot are calculated based on the coordinates for customers $s$, and $t$ distance is equal to the Euclidean distance of the dots $(x_s; y_s)$ and $(x_t; y_t)$. The time distances are equal to the path distances. Solomon created 6 different sets of instances: R1, R2, C1, C2, RC1 and RC2. In R-sets, the coordinates of the customers were chosen at random, from a defined interval, so the customers are evenly distributed everywhere. In C-sets the customers are clustered, which means that in several places, a greater number of customers is present. RC sets are something between the sets R and C, the customers are not evenly distributed, and they are not completely clustered either. In sets 1, the time windows were chosen from a smaller interval when compared to sets 2. That way, the number of customers per vehicle is significantly greater in sets 2, and with that, the number of routes is smaller. In Solomons' instances, the first goal of the optimization was the number of routes, then the total distance travelled by the vehicles. In this case, the smaller the number of routes is, the more optimal the solution is, and if two solutions have the same number of routes, the one with less distance travelled is optimal.

Table 1: Testing results for Solomons' instances

| Instance | Optimal solution (cost) [44] | Proposed ALGORITHM solution (cost) | From the optimal [%] | Number of vehicles optimal [44] | Number of vehicles ALGORITHM | Comments |
|---|---|---|---|---|---|---|
| c101 | 828,94 | 828,94 | 0 | 10 | 10 | All constraints met |
| c201 | 591,56 | 591,56 | 0 | 3 | 3 | All constraints met |
| r101 | 1650,8 | 1657,26 | 0,0646 | 19 | 18 | All constraints met, cost slightly greater but one vehicle less used. |
| r201 | 1252,37 | 1225,14 | -0,2723 | 4 | 5 | All constraints met, but a greater number of vehicles used. |
| rc101 | 1696,95 | 1637,62 | -0,5933 | 15 | 17 | All constraints met, but a greater number of vehicles used. |
| rc201 | 1406,94 | 1375,93 | -0,3101 | 4 | 5 | All constraints met, but a greater number of vehicles used. |
| c104 | 824,78 | 842,61 | 2,162 | 10 | 10 | All constraints met and an identical number of vehicles used. |
| r103 | 1292,68 | 1229,76 | -4,87 | 13 | 15 | All constraints met, but a greater number of vehicles used. |
| rc102 | 1554,75 | 1554,92 | -0,63 | 12 | 13 | All constraints met, but a greater number of vehicles used. |
| rc207 | 1061,14 | 1013,35 | -4,5 | 3 | 5 | All constraints met, but a greater number of vehicles used. |

Each of the 6 listed sets of instances contains 8-12 instances which are generated in the same way. For the algorithm testing, two have been taken from each group. Considering that the first criteria of Solomons' instances is the number of routes, and the main idea of the algorithm presented in this work is to minimize the cost while meeting all of the set constraints, for some instances the algorithm found a solution with optimal cost, but with a greater number of routes (Table 1).

Table 2: Results of testing on instances for 200 and 400 customers

| Instance | Optimal solution (cost) [44] | Proposed ALGORITHM solution (cost) | From the optimal [%] | Number of vehicles optimal [44] | Number of vehicles ALGORITHM | Comments |
|---|---|---|---|---|---|---|
| c1_2_1 | 2704,57 | 2704,57 | 0 | 20 | 20 | All constraints met |
| c2_2_1 | 1931,44 | 1983,82 | 0,5238 | 6 | 6 | All constraints met |
| r1_2_1 | 4784,11 | 4961,81 | 1,777 | 20 | 23 | All constraints met, but a greater number of vehicles used. |
| r2_2_1 | 4483,16 | 3827,98 | -6,5518 | 4 | 8 | All constraints met, but a greater number of vehicles used. |
| rc1_2_1 | 3602,8 | 3606,78 | 0,0398 | 20 | 23 | All constraints met, but a greater number of vehicles used. |
| rc2_2_1 | 3099,53 | 3169,49 | 0,6996 | 6 | 8 | All constraints met, but a greater number of vehicles used. |
| c1_4_1 | 7152,02 | 7152,29 | 0,0027 | 40 | 40 | All constraints met |
| c2_4_1 | 4116,05 | 4109,9 | -0,0615 | 12 | 15 | All constraints met, but a greater number of vehicles used. |
| r1_4_1 | 10372,31 | 10400,66 | 0,2835 | 40 | 40 | All constraints met |
| r2_4_1 | 9210,15 | 9456,51 | 2,675 | 8 | 9 | All constraints met, but a greater number of vehicles used. |
| rc1_4_1 | 8573,96 | 8580,71 | 0,0675 | 36 | 38 | All constraints met, but a greater number of vehicles used. |
| rc2_4_1 | 6682,37 | 6679,99 | -0,0238 | 11 | 12 | All constraints met, but a greater number of vehicles used. |

The algorithm was tested on Homberger and Gehring instances with 200 and instances with 400 customers. On instances with 200 customers, there was no significant difference when compared to Solomons' instances with 100 customers. Some instances have an improved solution in terms of cost, but the number of vehicles (routes) is greater compared to the optimal solution. From a practical viewpoint, the given solutions are optimal because they decrease the cost for the company. The optimality of the solution slightly decreases as the number of customers increases, but even for 400 customers (instance *rc2_4_1*), a better solution in terms of cost, and which meets all the given constraints, is obtained (Table 2).

The execution time of the implemented algorithm for Solomons' instances was up to a maximum of 1 [s]. The execution time was greater for instances with a greater number of customers, and so the execution of the algorithm for instances with 200 customers was up to a maximum of 5 [s], while for instances with 400 customers it was up to 200 [s]. One-third of the total execution time fell on the first step of the algorithm; the second step lasted for a negligibly short period of time, while the third step took up about 2/3 of the remaining execution time of the algorithm. The algorithm was executed on a laptop, with an Intel(R) Core(TM) i5-3210M CPU @ 2.50GHz 2.50 GHz processor, and 16 GB RAM DDR3 of memory.

## 4.2   Analysis of the results on real-world benchmark data

The proposed algorithm can meet constraints that are not defined in standardized data sets, as it is explained in great detail in previous sections. Because of that, testing of the given algorithm was done on real data from one of the biggest distribution company in Bosnia and Herzegovina. The data which was used for testing and which will be mentioned in the following text is placed on the 4TU.ResearchData [8] to be available as a new real-world benchmark dataset for the rest of the researchers.

Ten different days for which it was necessary to create the optimal transport routes, which meet all of the set constraints, were used for testing. Results are shown in Table 3. From the presented results, it can be concluded that in 9 out of 10 cases, all of the realistic constraints, which can be strictly defined and which can significantly aggravate the process of finding an optimal solution, are met. Only in one case, the constraint regarding the volume of one vehicle was violated, but the given overrun is equal to (0,004 / 3,15) * 100 = 0,13% of the permitted vehicle volume, which can be practically ignored. For the given day, the vehicles were filled up in terms of volume by 94,61%.

Another thing which can be concluded is that a small number of available vehicles (7 or 8), where most of the vehicles are different (different types and forms of vehicles, with different characteristics), which significantly aggravates the process of finding an optimal solution. The optimality of the solution primarily depends on the set constraints, and then on the number of customers and an available fleet of vehicles. Following that, it can be noticed that the solution cost with 94 customers is 1,5 greater than the solution cost for a route with 115 customers.

Table 3: Results of testing on real data of a distribution company

| Instance code [8] | Number of customers | ALGORITHM solution cost | Number of available vehicles | Number of vehicles used | Number of depots | Comments |
|---|---|---|---|---|---|---|
| 19062018 | 107 | 124,848 | 8 | 5 | 1 | All constraints met |
| 14062018 | 119 | 174,167 | 8 | 6 | 1 | All constraints met |
| 13062018 | 78 | 166,557 | 8 | 6 | 1 | All constraints met |
| 12062018 | 110 | 136,453 | 8 | 5 | 1 | All constraints met |
| 07062018 | 129 | 176,774 | 8 | 7 | 1 | All constraints met |
| 05062018 | 124 | 150,113 | 8 | 6 | 1 | All constraints met |
| 30052018 | 94 | 225,582 | 6 | 6 | 1 | All constraints met |
| 29052018 | 115 | 154,49 | 7 | 6 | 1 | All constraints met |
| 28052018 | 101 | 254,256 | 7 | 7 | 1 | Volume exceeded by 0,004 [m³] for vehicle A69-O-649! |
| 08052018 | 91 | 116,134 | 7 | 5 | 1 | All constraints met |

The execution of the algorithm is mainly affected by the number of customers. The total execution time of the algorithm ranges approximately from 300 to 500 [s]. It can be noticed that the given time is somewhat longer compared to the algorithm execution on standardized input data. That can be explained with a rather complex set of data with a great set of additional constraints. One-third of the total execution time fell on the first step of the algorithm, the second step lasted for a negligibly short period, while the third step took up about 2/3 of the remaining execution time of the algorithm. The algorithm was executed on a laptop, with an Intel(R) Core(TM) i5-3210M CPU @ 2.50GHz 2.50 GHz processor, and 16 GB RAM DDR3 of memory.

Table 4: Results of testing on real data of a distribution company – divided delivery

| Instance code [8] | Number of customers | ALGORITHM solution cost | Number of available vehicles | Number of vehicles used | Number of depots | Comments |
|---|---|---|---|---|---|---|
| 18012018 | 91 | 189,632 | 7 | 6* | 1 | All constraints met. (*) Because of the SD constraint vehicle J92-T-826, which can only service customer 139007, created two routes during the day, and the orders of customer 139007 are divided into two deliveries. In accordance with that, the working time of the vehicles are updated for vehicles which are found in more than one routing respectively. |

*An asterisk marks the fact that one vehicle made 2 deliveries (routes) in one day

The obtained results of the optimal routes were respected in real-life in a way that the drivers of the given distribution company successfully respected the presented routes, and managed to fully satisfy them. Given that the proposed algorithm supports the possibility of divided delivery for customers that cannot be serviced by one delivery (because of the set constraints), an actual input data set is available on the mentioned 4TU.ResearchData weblink [8], for which the algorithm calculated a cost shown in Table 4.

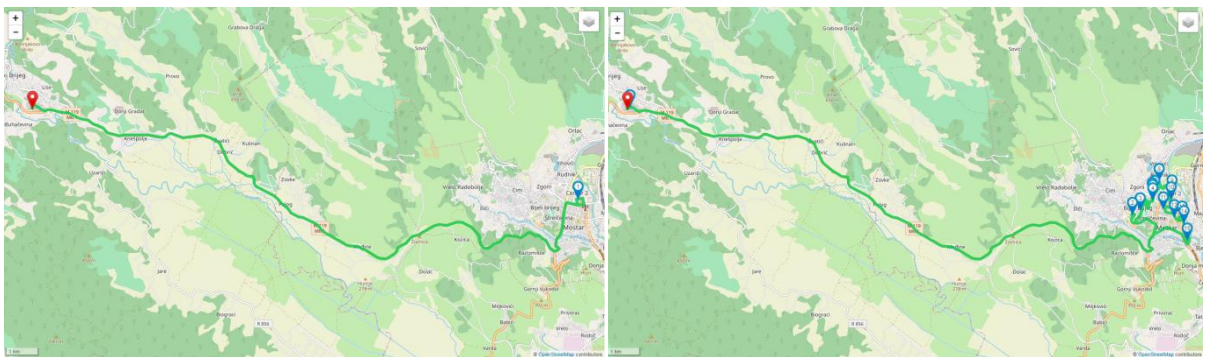

Figure 4: Example of divided delivery

The multiple deliveries shown are most noticeable on a graphic display for one of the used vehicles. On Figure 4, the left picture shows the first delivery and return of the vehicle to the depot, while the right picture shows another delivery for the given vehicle. The red colour marks the depot, while the customers have enumerated blue labels that mark the sequence of delivery for each of the customers.

Figure 5 shows an example of a typical delivery for one vehicle. The obtained optimal routes are shown on the map in order, where the order of servicing each customer is shown with a marker.

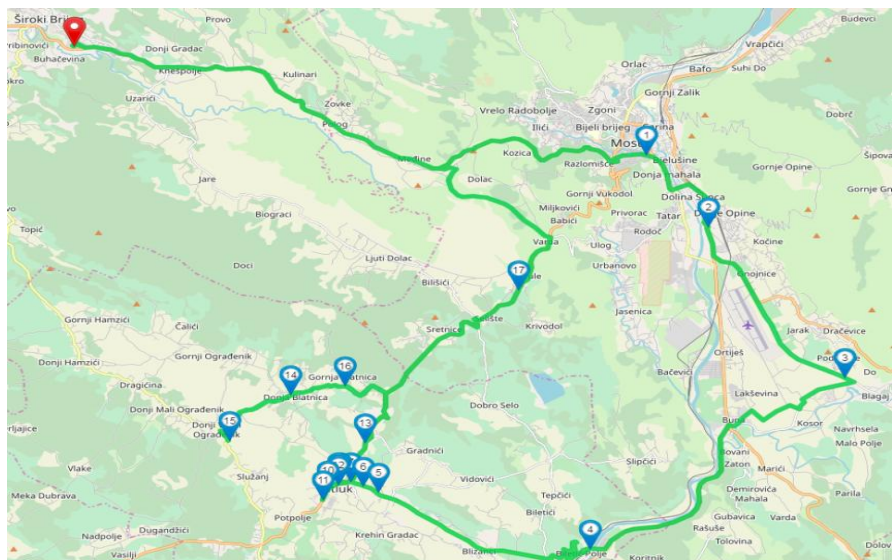

Figure 5: Example of a typical delivery for one vehicle

From the visual representation of the delivery, on the geographical map, it can be concluded that customers which have larger orders are serviced first, which is explained in greater detail in the introduction of Section 3.1.

## 4.3 Analysis results of control parameter adjustment of the proposed algorithm

As it was mentioned earlier, for every one of the control parameters, an independent regression model is created, with two regression algorithms: GLM and SVM. After obtaining the results, for each parameter, a Decision Support System, which determined the predicted value of the algorithm with *Predictive Confidence* based on the regression results, was created. To enrich the knowledge database of the control parameters, the algorithm was started 5632 times with all of the constraints met for a variety of different days and input parameters. The given data, which was used for testing and validation of the prediction model, is placed on the 4TU.ResearchData [9] to be available to the rest of the researchers.

Table 5: Comparative results of the used regression models

| Parameter [9] | GLM Predictive Confidence [%] | SVM Predictive Confidence [%] |
|---|---|---|
| $ToleranceWeight$ | 92,039 | 96,642 |
| $ToleranceVolume$ | 82,193 | 90,45 |
| $PenaltyDelay$ | 85,059 | 89,707 |
| $PenaltyCustomersVehicles$ | 92,044 | 96,647 |
| $CostIncreasing$ | 81,883 | 90,286 |
| $PenaltyPercentageVolume$ | 89,673 | 92,003 |
| $PenaltyPercentageWeight$ | 90,309 | 92,976 |

For each control parameter, a comparison of the *Predictive Confidence* [%] results was done, which, for the given input data set, is shown in Table 5. From the presented results it can be concluded that better prediction results were given for every SVM control parameter than for the GLM algorithm, and that is why the implemented *Decision Support System* preferred the prediction results of the SVM algorithm.

The obtained *Attribute Importance* segment in the prediction model helps determine the value of every one of the input attributes for the goal control parameter. The average value of the input parameters for the output control variables, as well as their order, based on the given value, is shown in Table 6.

Table 6: Attribute importance results

| Parameter [9] | Rank | Importance |
|---|---|---|
| $NUMBER\_VEHICLE\_TYPES$ | 1 | 0,871 |
| $NUMBER\_AVAILABLE\_VEHICLES$ | 2 | 0,861 |
| $SUM\_TIME\_WINDOWS\_TOTAL$ | 3 | 0,796 |
| $NUMBER\_OF\_ARTICLES\_TOTAL$ | 4 | 0,758 |
| $NUMBER\_OF\_CUSTOMERS\_TOTAL$ | 5 | 0,747 |
| $WEIGHT\_TOTAL$ | 6 | 0,741 |
| $NUMBER\_OF\_DIFFERENT\_CITIES$ | 7 | 0,727 |
| $VOLUME\_TOTAL$ | 8 | 0,582 |
| $NUM\_CONST\_CUST\_VEH\_TOTAL$ | 9 | 0,303 |

Analysis of the average values of the effect the input attributes have on every one of the order parameters concludes that the input parameters affect the resulting prediction control parameters in the order presented in Table 6. The achieved results are as expected because the routing was exceptionally complex with strict constraints when using actual data with a smaller number of available vehicles. This was especially affected by the fact that an average of 8 vehicles was available for routing, and seven of those were of different types and varieties, which significantly affected the results and the complexity of the algorithm execution. Based on that, the conclusion arises that those parameters are the most important ones in adjusting the control parameters of the algorithm. Also, what can be concluded is that the time windows of customers are of great importance to control parameters, which significantly affects the complexity of the solution search.

The parameter which, according to Table 6, has the least significance in adjusting the values of control parameters is the number of constraints, in terms of which customer cannot be serviced by which vehicle. The given number is presented in the form of a summarized indicator. If it were presented in the form of a ratio, of the customer to the number of vehicles which can service that customer, then the significance of that parameter would increase, and it could be the most important one.

For every one of the control parameters, the given results are presented graphically as well. For example, results for the parameter *ToleranceWeight* are shown in Figure 6.

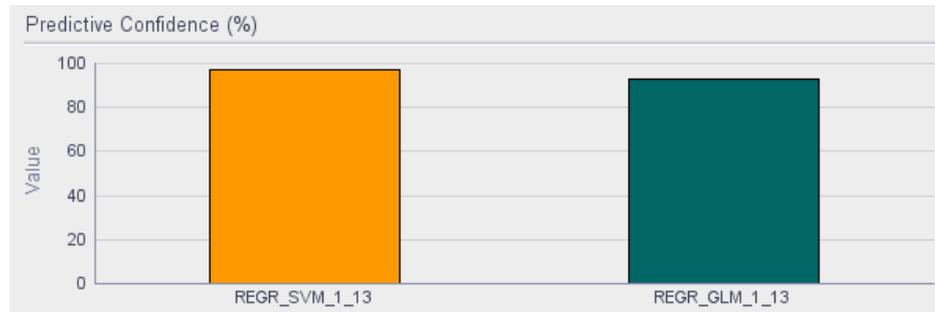

Figure 6: *Predictive Confidence* [%] comparison results for one control parameter

Also, it is possible to observe the comparison of Residual (Residual is the difference between the expected and predicted value of the dependent variable) for every one of the control parameters. An example of the comparison for the parameter *ToleranceWeight* is shown in Figure 7.

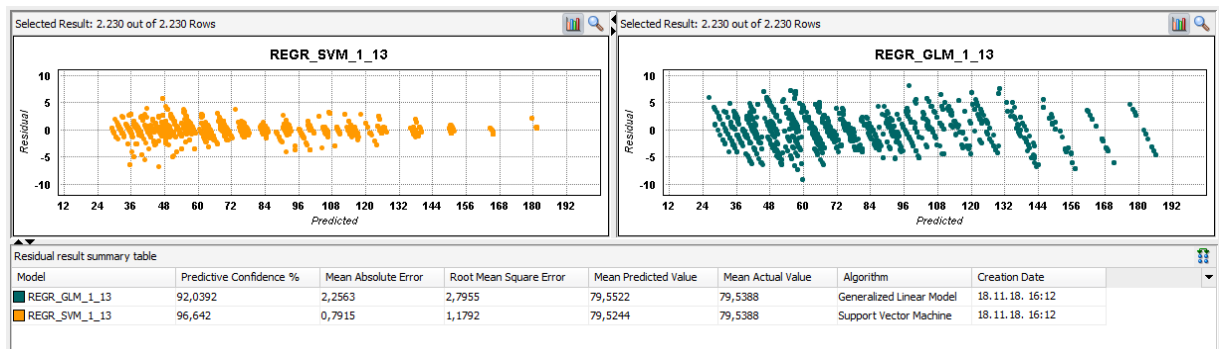

Figure 7: Residual comparison

The previous picture shows that for each attribute, except for the mentioned *Predictive Confidence* indicator, it is possible to obtain a multitude of other parameters (they primarily refer to the values of prediction errors) based on which it is possible to perform other comparisons and choose the model which meets all of the expectations and needs.

## 4.4  Practical significance of the proposed approach

No matter which of the approaches and methods are used for solving VRP in real conditions, there is always a risk that the given routes are not entirely feasible. Primarily this depends on the accuracy of the set parameters, input data and limitations. This work aims to present how some of the basic data required for solving and applying VRP can be set in a real environment. Practically feasible routes are the only ones that companies are interested in. In case the routes they get are not completely implemented a new level of insecurity in the operation of the implemented algorithm and model arises.

Table 7. Practical significance of the proposed approach (comparison results)

| | Average number of used vehicles | Average distance [km] | Percentage of route feasibility in realistic environment |
|---|---|---|---|
| BEFORE introducing the proposed approach | 55/10 = 5,5 | 129,761 | 79,65% |
| AFTER introducing the proposed approach | 57/10 = 5,7 | 137,915 | 99,14% |

As a practical conclusion of all the presented results, it can be concluded that the proposed approach has always tried to use smaller vehicles, while the bigger ones (with the much higher cost, for example, one of the biggest and most expensive vehicle in the dataset [8] labelled 875-M-523) were only included when necessary, or rather when it was not possible to serve all the customers using smaller vehicles. Even in this case, the bigger vehicle was used for serving fewer customers, with relatively smaller distances from the depot, to minimize its cost.

According to the comparative results presented in Table 7, it can be concluded that the feasibility of the routes has significantly increased (about 20%) by introducing the previously described approach. On the other hand, two vehicles more were used in 10 days during the routing, and the routing costs, as well as the total distance, have been slightly increased. However, the feasibility of the routes in realistic circumstances is the most important criterion. Even if the routes are optimal but not feasible, the transport processes of the company are much more difficult, so they increase the costs because there are some customers not being served as it was planned. On the other hand, it distorts one important factor, the company reputation, even more, which can result in a decreased number of satisfied customers, cancellations of the principals, and finally, it can result in the closure of the company itself. Therefore, this work takes on a strategic epithet and is crucial for every company that transports the goods.

# 5 Conclusions and Future research

This work presents a complex vehicle routing problem in the field of logistics with time windows and a set of real constraints, as well as a modular algorithm which adaptively solves that problem. The proposed algorithm consists of four steps. In the first step, an initial solution to the problem is created. It uses a modification of the heuristic Clarke-Wright algorithm. After that, the second intermediate step strives to decrease the number of routes. The result (routes) of the second step serves as an initial solution for the local Tabu search, which is the third step. After Tabu search, the fourth step (post-step) strives to optimize the sequence of customers within each of the previous routes. Before that, to significantly simplify the problem, the transformation of input data (time windows of customers and time distances) is performed. This enables the delivery time at each customer to become 0. Besides that, warehouse and vehicle working times are transformed in the pre-step, as well as a delivery division in the cases where one customer cannot be serviced by a certain vehicle only once.

The proposed algorithm consists of constants and control parameters, which are determined in a unique way using the knowledge base from historical data, based on the Generalized Linear Models and Support Vector Machine regression model. Both models make up the inputs in the Decision Support System, whose main task is to determine the best values for each of the algorithm's attributes, using the previously mentioned regression models. The stated procedure of the approximation of the best values of the control parameters for the given input data set is done in the phase of data preparation for each routing. The given procedure of pre-processing the algorithm's parameters gives an adaptive character to the whole approach.

The presented modular, adaptive approach can solve real-life VRP problems with several hundred delivery locations while meeting all of the set real-life constraints. Testing of the algorithm was done in two phases. The first testing was done using standardized data sets, where the implemented algorithm showed highly satisfying results. For some input data sets the proposed algorithm produced better results, up to 6,5% compared to the currently existing optimal solutions. For other routings it produced slightly less good results (never more than 3% than the optimal) compared to the current optimal solution of the given instances. The second type of testing was done on an actual data set, which was also published on the web link to serve other scientists in their researches and comparisons. For those data sets, the algorithm mostly managed to meet all of the very strict set constraints, and despite the minimal cost, the execution time of the algorithm was satisfactory from the aspects of a real-life application. The algorithm also has the ability of adaptability through automatic self-adjustment of the control parameters so that it is better and more advanced with every routing.

The approach and this algorithm is in use in some of the biggest distribution companies, as an implemented web-based enterprise system. The system enables human-computer interaction by allowing the subsequent manual modification of the obtained transport routes. It enables a graphic representation of the routes and comparison of the results. Results also showed that proposed improvements are increased by approximately 20% in the execution of the obtained routes in a real environment. In this way, the company is assured of the quality of the generated transportation routes and their customers have confidence in the delivery.

Based on the detailed description of the individual sections, it can be concluded that the contribution of this work is reflected in several ways. The basic contribution is based on the proposal of a modular, data-driven approach for successfully solving the vehicle routing problem that can be applied to the real cases in the field of logistics. An innovative, predictive and adaptive method of setting up and adjusting the parameters and constants used in the implementation of VRP algorithms which is based on the historical data is presented. The proposed approach, aside from the prediction model for the used parameters and constants also consists of the multi-step algorithm that is able to solve complex real-world VRP problems, accepting some of the constraints and facts that are not taken into consideration in other scientific papers in this scientific field. These are essential for the practical usage, feasibility and cost effectiveness of the resulting transportation routes in a realistic environment. These innovative segments of the proposed algorithm are also one of the contributions to this work, which are emphasized in a detailed description of the algorithm itself. In addition, the contribution of this work is also reflected in the fact that the real benchmark dataset is published to other researchers for further analyses and experiments. A practical contribution is reflected in the implementation of the web-based, easy-to-use system based on the proposed approach, with the possibility of the subsequent modification of the obtained transportation routes included. The system is being used by several of the largest distribution companies in Bosnia and Herzegovina. The resulting transportation routes are completely feasible, which results in financial and other benefits of the very companies using it.

Future research can be based on using Variable Neighborhood Search (VNS) or Simulated Annealing for the third step of the algorithm. Also, the third step could use a hybrid approach, which combines multiple metaheuristic algorithms during the local search. Another improvement could be decreasing the run-time of the algorithm's operators. One idea used in several of the mentioned scientific works relies on the approach of not observing all the possible combinations and choosing the best one, but rather only observing those combinations in which two geographically close routes can be swapped. Besides geographical location for certain parameter adjustments GPS data, weather conditions and delivery times can be used.

## Data Availability

The data used to support the findings of this study have been deposited in the online repositories:

[1] https://doi.org/10.4121/uuid:598b19d1-df64-493e-991a-d8d655dac3ea

[2] https://doi.org/10.4121/uuid:97006624-d6a3-4a29-bffa-e8daf60699d8

## Conflicts of Interests

The authors declare that they have no conflicts of interest.

## Acknowledgments

The authors would like to thank the Faculty of Electrical Engineering in Sarajevo for the resource support, and Info Studio d.o.o. Sarajevo for the possibility of practical use and testing.